\documentclass[10pt,twocolumn,letterpaper]{article}

\usepackage{iccv}
\usepackage{times}
\usepackage{epsfig}
\usepackage{graphicx}
\usepackage{mathtools}
\usepackage{tikz,pgfplots}
\usepackage{amsmath,bm,amssymb}
\usepackage{arydshln}
\usepackage{xurl}

\usepackage[pagebackref=true,breaklinks=true,letterpaper=true,colorlinks,bookmarks=false]{hyperref}

\iccvfinalcopy 

\def\iccvPaperID{2001} 

\begin{document}

\title{Probabilistic Monocular 3D Human Pose Estimation with Normalizing Flows}

\author{Tom Wehrbein\textsuperscript{1}
\and
Marco Rudolph\textsuperscript{1}
\and
Bodo Rosenhahn\textsuperscript{1}
\and
Bastian Wandt\textsuperscript{2}
\and
\textsuperscript{1}Leibniz University Hannover, \textsuperscript{2}University of British Columbia\\
{\tt\small wehrbein@tnt.uni-hannover.de}
}

\maketitle


\renewcommand{\footnotesize}{\fontsize{7.4pt}{9pt}\selectfont}
\begin{abstract}
3D human pose estimation from monocular images is a highly ill-posed problem due to depth ambiguities and occlusions.
Nonetheless, most existing works ignore these ambiguities and only estimate a single solution.
In contrast, we generate a diverse set of hypotheses that represents the full posterior distribution of feasible 3D poses.
To this end, we propose a normalizing flow based method that exploits the
deterministic 3D-to-2D mapping to solve the ambiguous inverse 2D-to-3D problem. 
Additionally, uncertain detections and occlusions are effectively modeled by incorporating uncertainty information of the 2D detector as condition.
Further keys to success are a learned 3D pose prior and a generalization of the best-of-M loss.
We evaluate our approach on the two benchmark datasets Human3.6M and MPI-INF-3DHP, outperforming all comparable methods in most metrics. The implementation is available on GitHub\footnote{\url{https://github.com/twehrbein/Probabilistic-Monocular-3D-Human-Pose-Estimation-with-Normalizing-Flows}}.

\end{abstract}

\renewcommand{\footnotesize}{\fontsize{8pt}{9pt}\selectfont}
\section{Introduction}
\begin{figure}
\begin{center}
\includegraphics[width=0.80\linewidth]{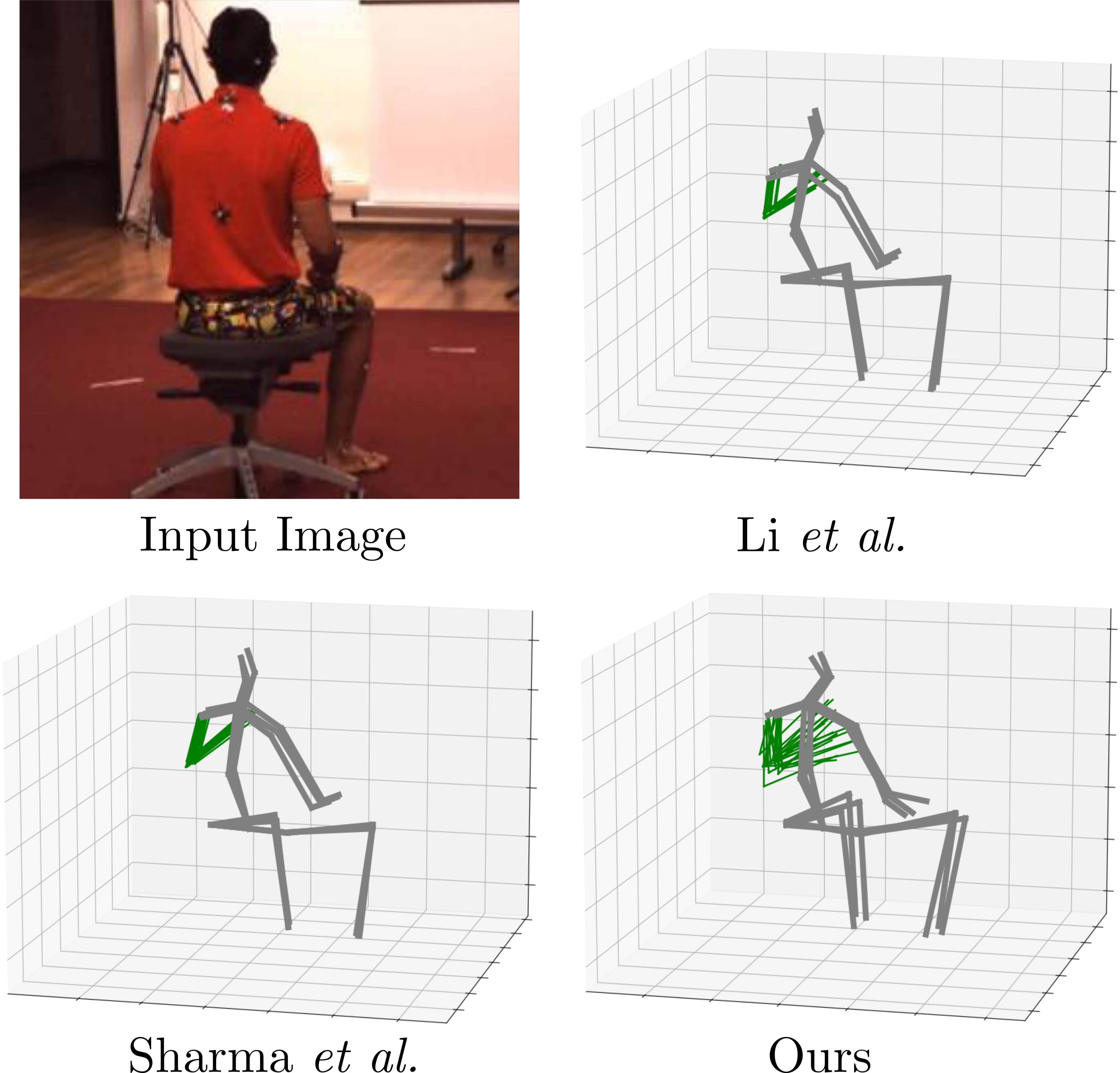}
\caption{Our model generates diverse 3D pose hypotheses that are consistent with the input image.
Compared to \cite{Li_2019_CVPR,Sharma_2019_ICCV} we achieve a higher diversity mainly where 2D detections are uncertain, in this case for the occluded left arm.
For visualization purposes, more than three hypotheses are shown only for the highly ambiguous left arm.}
\vspace{-1.5em}
\label{fig:teaser}
\end{center}
\end{figure}

Estimating the 3D pose of a human from a single monocular image is an active research field in computer vision.
It has many applications \eg in human computer interaction, animation, medicine and surveillance. 
A common approach is to decouple the problem into two stages. 
In the first stage, a 2D pose detector is used to estimate 2D keypoints which are then lifted to 3D joint locations in the second stage.
By utilizing a 2D pose detector pretrained on diverse and richly annotated data,
the 3D pose estimator becomes invariant to different scenes varying in lighting, background and clothing.
However, reconstructing the correct 3D pose from 2D joint detections is a highly ill-posed problem because of depth ambiguities and occluded body parts.
While some ambiguities can be resolved by utilizing information from the image (\eg difference in shading due to depth disparity) or by exploiting known proportions of the human body, such as joint angle and bone length constraints, there still remain scenarios where multiple plausible 3D poses are consistent with the same image.
Fig.~\ref{fig:teaser} shows such a situation where the left arm is occluded by the upper body and therefore its position cannot be determined unambiguously.
Nevertheless, most existing works ignore the ambiguities by assuming
that only a single solution exists.
In contrast, we model monocular 3D human pose estimation as an ambiguous inverse problem with multiple feasible solutions.
Thus, in this work, we propose to estimate the full posterior distribution of plausible 3D poses conditioned on a monocular image.

Recently, few methods \cite{Jahangiri2017GeneratingMD,Li_2019_CVPR,li2020weakly,oikarinen2020graphmdn,Sharma_2019_ICCV} have been proposed that follow the line of research to explicitly generate multiple 3D pose hypotheses from the 2D input.
However, they only consider 2D joint coordinates and ignore the uncertainty of the 2D detector.
While it is reasonable to infer depth ambiguities based on 2D coordinates only, 
directly modeling occlusions and uncertain detections is not meaningful.
Fortunately, most 2D human joint detectors encode valuable information about uncertainties of the location of human joints in the predicted heatmaps.
Instead of discarding this information, we propose to explicitly extract and utilize the uncertainties of the 2D detector from the estimated heatmaps.
As shown in Fig.~\ref{fig:teaser}, this enables us to effectively model the uncertainties of the 2D detector together with the inherent depth ambiguities.

In this work, we propose a normalizing flow based method inspired by the framework for solving ambiguous inverse problems from Ardizzone \etal~\cite{ArdizzoneINN}.
A normalizing flow \cite{pmlr-v37-rezende15,doi:10.1002/cpa.21423,tabak2010} is a sequence of bijective transformations which allows evaluation in both directions.
We propose to view 3D human pose estimation from a single image as an ambiguous inverse problem, since it is a deterministic forward process (\ie projection of the 3D pose to 2D) with multiple different inverse mappings.
Constructing a bijection between a 3D pose and the combination of a 2D pose with a latent vector allows to utilize the 3D-to-2D mapping (forward process) during training.
Intuitively, depth information that otherwise gets lost in the forward process is encoded in the latent vector.
Repeatedly sampling the latent vector and
computing the inverse path of the normalizing flow generates
arbitrary many 3D pose hypotheses that approximate the true posterior distribution.
To incorporate the uncertainty information from the heatmaps, we employ a conditional variant of normalizing flows \cite{Ardizzone2019a,winkler2019learning}.
We extract the uncertainty information by fitting 2D Gaussians to the heatmaps
which are then used to form a conditioning vector.
We optimize the model in both directions.
The forward path learns the 3D-to-2D mapping and to produce latent vectors following a predefined distribution. 
For the inverse path, we utilize the 3D pose discriminator of \cite{Wandt2019RepNet} to penalize anthropometrically unfeasible poses.
Additionally, we apply a loss enforcing the 3D pose hypotheses to reflect the uncertainties of the 2D detector.
Motivated by common practice in particle filters, we further propose a generalization of the \textit{best-of-M} loss \cite{NIPS2012_cfbce4c1} that minimizes the distance between the mean of the $k$ best hypotheses and the corresponding ground truth.

We evaluate our approach on the two benchmark datasets Human3.6M \cite{h36m_pami} and MPI-INF-3DHP \cite{mono-3dhp2017} and outperform all comparable methods in most metrics.
Given the focus on ambiguous examples, we further evaluate on a subset of Human3.6M containing only samples with a high degree of 2D detector uncertainty.
On this subset, our method outperforms the competitors by a large margin.
To summarize, our contributions are:
\begin{itemize}
    \item To the best of our knowledge, we are the first to employ a normalizing flow based method for modeling the posterior distribution of 3D poses given a single image.
    \item Uncertainty information from the predicted heatmaps of the 2D detector is incorporated into our method, enabling to effectively model occlusions and uncertain detections.
    \item We propose a generalization of the \textit{best-of-M} loss that noticeably improves prediction performance.
\end{itemize}

\section{Related Work}
In this section, we first give an overview of recent work in 3D human pose estimation focused on two-stage approaches. 
Afterwards, existing methods for multi-hypotheses 3D pose generation are discussed, followed by an overview of relevant work on normalizing flows.
While there has been recent interest in estimating the 3D human body shape from monocular images \cite{multi_bodies_biggs2020,KanazawaCVPR18,kocabas2019vibe,kolotouros2019spin,li21hybrik,pavlaCVPR18,xu2020ghum,Zanfir2020WeaklyS3},
this work focuses on predicting the 3D locations of a set of predefined joints.

\textbf{Lifting 2D to 3D:}
Our approach belongs to the vast body of work that estimate 3D poses from the output of a 2D pose detector
\cite{Chen_2017_CVPR,Ci_2019_ICCV,fang2018learning,inthewild3d_2019,Hossain2018ECCV,Li_2020_CVPR,PhysCapTOG2020,Wandt2019RepNet,WanRud2021a,DBLP:journals/corr/abs-1905-07862,Xu_2020_CVPR}. These two-stage approaches decouple the difficult problem of 3D depth estimation from the easier 2D pose localization. Furthermore, it allows to use both indoor and in-the-wild data for training the
2D detector, which effectively reduces the bias towards sterile indoor scenes. 
Akhter and Black~\cite{Akhter_2015_CVPR} learn a pose-conditioned joint angle limit prior to restrict invalid 3D pose reconstructions. They perform 3D pose estimation using an over-complete dictionary of poses. 
Moreno-Noguer~\cite{Moreno-Noguer_2017_CVPR} casts the problem as a regression between 2D and 3D poses represented as distance matrices.
Lifting 2D to 3D joints was further sparked by Martinez \etal~\cite{martinez_2017_3dbaseline}, who employ a simple fully-connected network to lift 2D detections to 3D poses, surprisingly outperforming past approaches.
Due to its simplicity and strong performance, it serves as a popular baseline for many following works.

Unlike the above mentioned approaches that assume a unimodal posterior distribution and only predict a single 3D pose for each input, we are able to generate a diverse set of plausible 3D poses.
Additionally, anatomical constraints are learned implicitly by utilizing a strong 3D pose discriminator.
In contrast to previous works that integrate uncertainty information of the 2D detector (\eg \cite{BogoECCV2016,WanRud2021a,Xu_2020_CVPR}), we fit a 2D Gaussian to each heatmap instead of 
using only the maximum value of each heatmap as confidence score,
thus better capturing the uncertainty distribution.
\begin{figure*}[!ht]
\begin{center}
\includegraphics[width=1.0\linewidth]{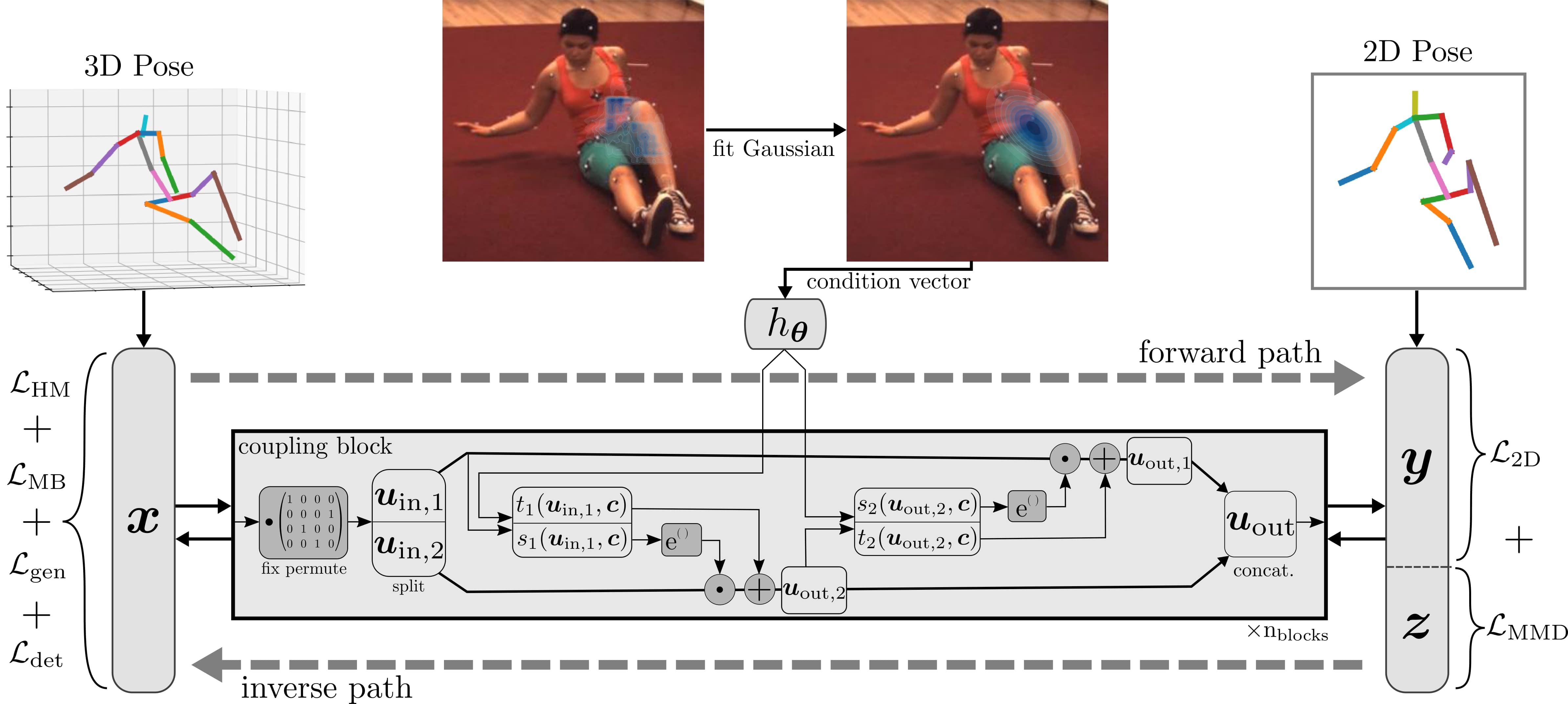}
\caption{An overview of our proposed method. 
We employ a normalizing flow consisting of affine coupling blocks \cite{DBLP:conf/iclr/DinhSB17} for generating multiple 3D pose hypotheses.
By constructing a bijection between a 3D pose and the concatenation of a 2D pose with a latent vector, we can exploit the 3D-to-2D mapping (forward path) during training.
The model is optimized in both directions, whereas at inference, only the path from 2D to 3D (inverse path) is computed.
Arbitrary many 3D pose hypotheses can be generated by repeatedly sampling the latent vector from a known distribution and computing the inverse path.
Uncertainty information of the 2D detector in form of fitted Gaussians is incorporated by conditioning the coupling blocks. 
The architecture of a single coupling block is visualized in the gray box.
For visualization purposes, only the forward computation of the coupling block is shown. }
\vspace{-1.5em}
\label{fig:liftingINN}
\end{center}
\end{figure*}

\textbf{Multi-Hypotheses 3D Human Pose Estimation:}
There are early works \cite{LeeCo04,Simo12,SmiTri01,SmiTri03} that extensively analyze and discuss the ambiguities of monocular 3D human pose estimation and sample multiple 3D poses via heuristics.
More recently, Jahangiri and Yuille~\cite{Jahangiri2017GeneratingMD} propose to generate multiple hypotheses from a predicted seed 3D pose by uniformly sampling from learned occupancy matrices \cite{Akhter_2015_CVPR}.
Furthermore, they impose bone length constraints and reject hypotheses with 2D reprojection error larger than some threshold. 
Li and Lee~\cite{Li_2019_CVPR} employ a mixture density network (MDN) \cite{370fbeadb5584ba9ab2938431fc4f140} to learn the multimodal posterior distribution.
The conditional mean of each Gaussian kernel then denotes one 3D pose hypothesis.
Oikarinen \etal~\cite{oikarinen2020graphmdn} utilize the semantic graph neural network of \cite{zhaoCVPR19semantic} to improve upon the MDN approach of \cite{Li_2019_CVPR}.
Contrary to our normalizing flow based approach, the number of generated hypotheses needs to be specified a priori and is fixed for every input.
Furthermore, when increasing the number of generated hypotheses, significantly more computational resources are required. Sharma \etal~\cite{Sharma_2019_ICCV} employ a conditional variational autoencoder to synthesize diverse 3D pose hypotheses conditioned on a 2D pose detection.
They also propose to derive joint-ordinal depth relations from the image to rank the 3D pose samples. 
In contrast to \cite{Jahangiri2017GeneratingMD}, our normalizing flow based approach does not need to incorporate computationally heavy rejection sampling or requires to define the number of generated 3D pose hypotheses a priori.
Our method is more flexible and is able to model any posterior distribution without requiring explicit hard constraints. 
Moreover, we are the only ones to incorporate the uncertainty information of the 2D detector, enabling us to significantly improve on highly ambiguous cases and to inherently handle an arbitrary number of occluded joints.

\textbf{Normalizing Flows:}
A normalizing flow \cite{pmlr-v37-rezende15,doi:10.1002/cpa.21423,tabak2010} is a sequence of bijective transformations that transforms a simple tractable distribution into a complex target data distribution.
Because of the bijectivity, evaluation in both directions is possible.
Namely, sampling data from the modeled distribution as well as exact density estimation (\ie assigning a likelihood to each data point).
Most common state-of-the-art flow architectures are based on auto-regressive models that utilize the Bayesian chain rule to decompose the density \cite{DBLP:journals/corr/DinhKB14,DBLP:conf/iclr/DinhSB17,pmlr-v37-germain15,NIPS2016_6581,NIPS2017_6828,RudWan2021}.
For a more comprehensive introduction, we refer the reader to \cite{kob2020NF}.

Ardizonne \etal~\cite{ArdizzoneINN} extend the real-valued non-volume preserving (Real-NVP) transformations from Dinh \etal~\cite{DBLP:conf/iclr/DinhSB17} to the task of computing posteriors for ambiguous inverse problems.
Given such an ambiguous inverse problem, they propose to learn the well-understood forward process in a supervised manner and encode otherwise lost information in additional latent variables.
Thus, they learn a bijective mapping between the target data distribution and the joint distribution of latent variables and forward process solutions.
Due to invertibility, the inverse is implicitly learned.
By repeatedly sampling the latent variables from a simple tractable distribution, they can approximate the full posterior. Inspired by their work, we adopt and extent their framework for modeling the full posterior distribution of plausible 3D poses conditioned on a monocular image.
We introduce a conditioning vector, a learnable prior and two additional loss functions.

To the best of our knowledge, the only previous works in human pose estimation utilizing normalizing flows are \cite{multi_bodies_biggs2020,xu2020ghum,Zanfir2020WeaklyS3}. However, they employ normalizing flows as 3D pose prior and not for directly modeling the posterior distribution of 3D poses conditioned on an image.

\section{Method}
Our aim is to learn the full posterior distribution of plausible 3D poses conditioned on a monocular image. We follow the popular two-stage approach by first applying a state-of-the-art 2D joint detector \cite{Sun_2019_CVPR} and subsequently using its output to estimate corresponding 3D pose hypotheses. The core idea is that instead of conditioning the posterior distribution only on the 2D detections, we additionally utilize uncertainty information extracted from the predicted heatmaps in a novel way. This enables to effectively model the uncertainties of the 2D detector together with the inherent depth ambiguities.

An overview of the proposed method is shown in Fig.~\ref{fig:liftingINN}.
To learn the posterior distribution, we employ a normalizing flow to construct a bijective mapping between a 3D pose $\bm{x} \in \mathbb{R}^{3J}$ and the concatenation of a 2D pose $\bm{y} \in \mathbb{R}^{2J}$ with a latent vector $\bm{z} \in \mathbb{R}^{J}$, where $J$ is the number of joints in one pose.
The introduction of the latent vector $\bm{z}$ allows to utilize the well-defined forward process of projecting a 3D pose to its 2D observation during training.
Intuitively, $\bm{z}$ captures depth information that is otherwise lost in the mapping from 3D to 2D.
Instead of simply using the argmax of the heatmaps, we incorporate the uncertainty information of the 2D detector by conditioning the normalizing flow on Gaussians fitted to the heatmaps.
At inference, the full posterior is approximated by repeatedly sampling $\bm{z}$ from the distribution of latent variables and computing the inverse path.
If the forward process is simulated successfully, all generated hypotheses reproject to the corresponding 2D pose observation.

\subsection{Conditional Normalizing Flow}
As normalizing flow we adopt the Real-NVP \cite{DBLP:conf/iclr/DinhSB17} affine coupling block architecture.
This architecture can straight-forwardly be extended to incorporate a conditional input \cite{Ardizzone2019a,winkler2019learning}.
A single coupling block is shown in the gray box in Fig.~\ref{fig:liftingINN}. The input $\bm{u}_{\mathrm{in}}$ is split into two parts $\bm{u}_{\mathrm{in}, 1}$ and $\bm{u}_{\mathrm{in}, 2}$. Subsequently, $\bm{u}_{\mathrm{in}, 1}$ and $\bm{u}_{\mathrm{in}, 2}$ undergo a scale and translation transformation parameterized by the functions $s_i$ and $t_i$ ($i \in \{1, 2\}$) on two separate paths. The outputs $\bm{u}_{\mathrm{\mathrm{out}}, 1}$ and $\bm{u}_{\mathrm{\mathrm{out}}, 2}$ are concatenated to form the overall output of the coupling block. Given the heatmap condition $\hat{\bm{c}}$, further encoded into the conditioning vector $\bm{c} = h_{\bm{\theta}}(\hat{\bm{c}})$, the forward path of a coupling block is defined as
\begin{equation}
 \begin{aligned}
\bm{u}_{\mathrm{\mathrm{out}}, 2} = \bm{u}_{\mathrm{in}, 2} \odot e^{s_1(\bm{u}_{\mathrm{in}, 1}, \bm{c})} + t_1(\bm{u}_{\mathrm{in}, 1} , \bm{c}) \\
\bm{u}_{\mathrm{\mathrm{out}}, 1} = \bm{u}_{\mathrm{in}, 1} \odot e^{s_2(\bm{u}_{\mathrm{\mathrm{out}}, 2}, \bm{c})} + t_2(\bm{u}_{\mathrm{\mathrm{out}}, 2}, \bm{c}),
\end{aligned}
\label{aff}
\end{equation}
where $\odot$ denotes the element-wise multiplication. The exponential function is used to prevent multiplication by zero, which ensures the invertibility of the block. Note that $s_i$ and $t_i$ represent functions that do not need to be invertible. The only restriction is that their produced output matches the dimensions of the data on the corresponding path in the coupling block. Instead of regressing the scale and translation coefficients separately, we employ a fully-connected network that jointly predicts them by splitting its output. 
By construction, the coupling block can be trivially inverted without any computational overhead.
The overall network consists of multiple chained blocks, each followed by a predefined random permutation which shuffles the path assignment of the variables.
The output of the last block is split to form the 2D pose $\bm{y}$ and the latent vector $\bm{z}$.

Following Ardizzone \etal~\cite{Ardizzone2019a}, we adopt a parameterized soft clamping mechanism to prevent instabilities caused by the exponential function in the coupling block. The soft clamping is defined as
\begin{equation}\label{eq:softclamp}
\sigma_{\alpha}(r) = \frac{2\alpha}{\pi}\arctan{\frac{r}{\alpha}},
\end{equation}
and is applied as the last layer of $s_1$ and $s_2$. It prevents scaling components of exploding magnitude by restricting the output to the interval $(-\alpha, \alpha)$.

\subsection{Heatmap Condition}\label{sec:heatmapCond}
Recent 2D detectors are optimized by applying a supervised loss between the predicted heatmap and a ground-truth heatmap consisting of a 2D Gaussian centered at the joint location.
This leads to the predicted heatmaps being a valuable source of uncertainty of the 2D detector.
Instead of estimating 3D poses solely based on 2D joint coordinates, we incorporate the uncertainties of the 2D detector encoded in the estimated heatmaps.
Specifically, we fit a 2D Gaussian to each predicted heatmap to best capture the uncertainty distribution.
The fitting process is done using non-linear least squares.
As initial parameters, we set the amplitude to $1$, the mean of each Gaussian to the corresponding regressed 2D joint location and the covariance matrix to a diagonal matrix with $\sigma^2_x = \sigma^2_y = \sigma^2_{\text{gt}}$, where $\sigma^2_{\text{gt}}$ is the ground-truth variance used for training the 2D detector.
For each image, the fitted coefficients are stacked to form a single vector.
We discard the Gaussian coefficients for the hip joints, since the typical alignment of the root joint of the 3D poses heavily reduces the possible variances in these joints.
Thus, the heatmap conditioning vector is denoted as $\hat{\bm{c}} \in \mathbb{R}^{6(J-3)}$.
We employ a fully-connected network as encoding network $h_{\bm{\theta}}$ that further encodes $\hat{\bm{c}}$ into $\bm{c} = h_{\bm{\theta}}(\hat{\bm{c}})$.
For the 3D pose hypotheses to best reflect the uncertainties of the 2D detector, we explicitly optimize the network to match the 2D Gaussian distributions in the $x$- and $y$-direction of the 3D hypotheses for each joint.
Let $\hat{\Sigma} \in \mathbb{R}^{2 \times  2}$ be the covariance matrix of a single joint estimated from the positions of that joint in the $L$ produced hypotheses.
Defining $\Sigma \in \mathbb{R}^{2 \times  2}$ as the covariance matrix of the fitted 2D Gaussian of the corresponding heatmap, we minimize a masked lower bound Root Mean Square Error (RMSE) between both covariance matrices:   
\begin{equation}
 \begin{aligned}
    \mathcal{L}_{\text{HM}}\big({\Sigma}, &\hat{\Sigma}\big) = m \cdot \Big(\max\big(0, \Sigma_{1, 1} - \hat{\Sigma}_{1, 1}\big)^2 \\
    &+ \max\big(0, \Sigma_{2, 2} - \hat{\Sigma}_{2, 2}\big)^2 + \big(\Sigma_{1, 2} - \hat{\Sigma}_{1, 2}\big)^2 \Big)^{\frac{1}{2}},
\end{aligned}
\end{equation}
where the masking scalar $m$ is defined as
\begin{equation} \label{eq:masking}
    m = \left\{
    \begin{array}{ll}
1 & \sqrt{\Sigma_{1,1}} > \sigma_t \lor \sqrt{\Sigma_{2,2}} > \sigma_t\\
0 & \text{otherwise}
\end{array}\right..
\end{equation}
Thus, the loss has no influence if the 2D detector is certain about the location of the specific joint, indicated by a fitted Gaussian with standard deviations smaller than the threshold $\sigma_t$. 
To not unnecessarily restrict the network, we only penalize the diagonal entries of the covariance matrices if they are smaller than the corresponding ground-truth values.

\subsection{Optimization}
\label{sec:optimization}
The core idea of the optimization procedure is to train the 3D-to-2D mapping (forward path) in a supervised manner, while the highly ambiguous 2D-to-3D mapping (inverse path) is learned implicitly due to the invertibility of the normalizing flow and supported by additional supervision with the inverse process. 
Each training iteration
consists of first calculating the forward path, followed by $L$ computations of the inverse path and two additional, one for the discriminator and one for the deterministic 3D reconstruction. The gradients from both directions are accumulated before performing a parameter update. Note that due to the Real-NVP coupling block architecture, both directions can be computed efficiently.

\textbf{Forward Path:}
In the forward process, the network predicts the corresponding 2D joint detections given a 3D pose.
This is optimized using the $L_1$ distance:
\begin{equation} \label{eq:Lyloss}
    \mathcal{L}_{\text{2D}} = \left\lVert \bm{y} - \bm{\hat{y}} \right\rVert_1,
\end{equation}
where $\bm{y}$ is the ground-truth and $\bm{\hat{y}}$ the estimated 2D observation.
The estimated latent variables are optimized to follow a zero-mean isotropic Gaussian $p_{\bm{Z}} = \mathcal{N}(0, I)$ and to be independent from the distribution of 2D observations $p_{\bm{Y}}$. 
Both properties are enforced by minimizing the Maximum Mean Discrepancy (MMD) \cite{JMLR:v13:gretton12a} between the joint distribution of network outputs $q(\bm{y}, \bm{z})$ and the product of marginal distributions $p_{\bm{Y}}$ and $p_{\bm{Z}}$.
Given samples $\hat{\bm{V}} = {\{\hat{\bm{v}}_i}\}^n_{i=1}$ drawn i.i.d.\ from $q(\bm{y}, \bm{z})$ and $\bm{V} = {\{\bm{v}_i}\}^n_{i=1}$ with $\bm{v}_i = [\bm{y}, \bm{z}]$ and $\bm{y} \sim p_{\bm{Y}}$, $\bm{z} \sim p_{\bm{Z}}$,
the unbiased estimator of the squared MMD with kernel $\varphi$ is
\begin{equation}
 \begin{aligned}
\mathcal{L}&_{\text{MMD}} = \text{MMD}^2_{u}(\bm{V}, \hat{\bm{V}})=\frac{1}{n(n-1)} \sum_{i \neq j}^{n} \varphi\left(\bm{v}_{i}, \bm{v}_{j}\right) \\ &+ \frac{1}{n(n-1)} \sum_{i \neq j}^{n} \varphi\left(\hat{\bm{v}}_{i}, \hat{\bm{v}}_{j}\right) - \frac{2}{n^2} \sum_{i,j=1}^{n} \varphi\left(\bm{v}_{i}, \hat{\bm{v}}_{j}\right).
\end{aligned}
\end{equation}
Following \cite{ArdizzoneINN}, we block the gradients of $\mathcal{L}_{\text{MMD}}$ with respect to $\bm{y}$ to prevent the predictions of $\bm{y}$ from deteriorating.

\textbf{Inverse Path:} 
Given a 2D pose $\bm{y}$, a latent vector $\bm{z}$ is drawn from the base distribution $p_{\bm{Z}}$ and concatenated to form the input $[\bm{y}, \bm{z}]$ of the inverse path. By repeatedly sampling $\bm{z} \sim p_{\bm{Z}}$, arbitrary many 3D pose hypotheses can be created.
Although $\mathcal{L}_{\text{2D}}$ and $\mathcal{L}_{\text{MMD}}$ are in theory sufficient to best approximate the true posterior distribution \cite{ArdizzoneINN}, we apply additional losses to the inverse path to improve convergence. 
To penalize geometrically unfeasible 3D pose hypotheses, we introduce a discriminator network and adopt the Improved Wasserstein GAN training procedure of \cite{NIPS2017_wgangp}. The inverse path acts as the generator by producing 3D poses that minimize the negated output of the discriminator. This loss is denoted as $\mathcal{L}_{\text{gen}}$.
The architecture of the discriminator is taken from \cite{Wandt2019RepNet}, including a Kinematic Chain Space layer \cite{WanAck2018a} encoding bone lengths and angular representations. 
Additionally, we generate a 3D pose for each 2D input by using the corresponding latent vector $\bm{z}_{\text{det}}$ produced in the forward path. Note that contrary to sampling a latent vector $\bm{z} \sim p_{\bm{Z}}$, when using the estimated latent vector $\bm{z}_{\text{det}}$, it is reasonable to apply a supervised loss $\mathcal{L}_{\text{det}}$ linking a 2D input to a single 3D pose, since the combination of a predicted latent vector and matching 2D pose detection should correspond to the single exact solution of the ambiguous inverse problem. We minimize the $L_1$ distance between the ground-truth 3D pose $\bm{x}$ and the estimated 3D pose $\hat{\bm{x}}_{\text{det}}$
\begin{equation} \label{eq:Ldet}
    \mathcal{L}_{\text{det}} = \left\lVert \bm{x} - \hat{\bm{x}}_{\text{det}} \right\rVert_1.
\end{equation}
To further guide the optimization process, we propose a generalization of the \textit{best-of-M} loss \cite{NIPS2012_cfbce4c1}. 
Given a set of 3D pose hypotheses $\bm{H} = {\{\hat{\bm{x}}_i}\}^L_{i=1}$ generated from the same 2D input, we select the subset $\bm{H}_{\text{topk}} \subseteq \bm{H}$ consisting of the $k$ pose hypotheses with the lowest Mean Per Joint Position Error (MPJPE) to the corresponding ground-truth pose $\bm{x}$.
We then minimize the $L_1$ distance between the ground-truth pose $\bm{x}$ and the mean of the $k$ best hypotheses:
\begin{equation}
    \mathcal{L}_{\text{MB}} = \left\lVert \bm{x} - \frac{\sum_{\hat{\bm{x}} \in \bm{H}_{\text{topk}}}\hat{\bm{x}}} {k}  \right\rVert_1.
\end{equation}

\textbf{Overall:}
In total, the objective function of our normalizing flow is
\begin{equation} \label{eq:innLoss}
 \begin{aligned}
\mathcal{L}_{\text{NF}} = &\mathcal{L}_{{\text{2D}}} + \mathcal{L}_{\text{gen}} + \lambda_{\text{MMD}} \mathcal{L}_{\text{MMD}} \\ &+ \lambda_{\text{det}} \mathcal{L}_{\text{det}} + \lambda_{\text{MB}} \mathcal{L}_{\text{MB}} + \lambda_{\text{HM}} \mathcal{L}_{\text{HM}},
 \end{aligned}
\end{equation}
where $\lambda_{\text{MMD}}$, $\lambda_{\text{det}}$, $\lambda_{\text{MB}}$ and $\lambda_{\text{HM}}$ represent the weights of the corresponding losses. The discriminator network is optimized to distinguish between the 3D poses produced by the normalizing flow and 3D poses from the training set by minimizing the WGAN-GP objective function \cite{NIPS2017_wgangp}.
The encoding network $h_{\bm{\theta}}$ is jointly optimized with the normalizing flow by propagating gradients from $\mathcal{L}_{\text{NF}}$ through $h_{\bm{\theta}}$.

\section{Experiments}
\subsection{Datasets and Evaluation Metrics}
\textbf{Human3.6M} \cite{h36m_pami} is the largest video pose dataset for 3D human pose estimation.
It features $7$ professional actors performing 15 different activities, such as \textit{Sitting}, \textit{Walking} and \textit{Smoking}.
For each frame, accurate 2D and 3D joint locations and camera parameters are provided.
We follow the standard protocols and evaluate on every $64^{\text{th}}$ frame of subjects 9 and 11.
Protocol 1 computes the Mean Per Joint Position Error (MPJPE) between the reconstructed and ground-truth 3D joint coordinates directly, whereas Protocol 2 first applies a rigid alignment between the poses (PMPJPE).
We additionally show results for the Correct Poses Score (CPS) metric proposed by \cite{WanRud2021a}.
It considers a pose as correct if and only if all joints have a Euclidean distance to the ground-truth below a threshold value $\vartheta$. CPS is then defined as the area under curve for $\vartheta \in [0\text{\,mm}, 300\text{\,mm}]$.
Instead of evaluating the reconstruction joint by joint, the CPS considers the whole pose.
Compared to other common metrics, it is better suited to detect wrongly estimated poses that could negatively influence downstream tasks.

\textbf{Human3.6M Ambiguous (H36MA):} To focus evaluation on highly ambiguous examples, we select a subset\footnote{Information about the exact composition of the subset can be found in the official GitHub repository.} of the Human3.6M test split according to the uncertainties of the 2D detector.
This subset only contains samples for which at least one fitted Gaussian has a standard deviation larger than 5\,px, which holds true for $6.4\%$ of all samples in the test split.
These samples are extremely challenging since the joint detector gives inaccurate or wrong results.

\textbf{MPI-INF-3DHP} (3DHP) \cite{mono-3dhp2017} is a 3D human pose dataset containing annotated images recorded in three different settings: studio with green screen, studio without green screen and outdoors.
We evaluate on the test split without utilizing the training data to assess the generalization capability of our network.
Following previous works, the Percentage of Correct Keypoints (PCK) under 150\,mm is adopted as the metric for 3DHP.

\subsection{Implementation Details}
\textbf{2D Detector:} We use the publicly available HRNet \cite{Sun_2019_CVPR} pretrained on MPII \cite{andriluka14cvpr} as our 2D joint detector and finetune it on Human3.6M. Target ground-truth heatmaps are created with $\sigma_{\text{gt}} = 2\text{\,px}$.

\textbf{Data Preprocessing:} We center each 2D pose to its mean and divide it by its standard deviation.
The 3D poses are processed in metres and also mean centered individually.
Before evaluation, 3D poses are zero-centered around the hip joint to follow the standard protocols.

\textbf{Network Details:}
The normalizing flow consists of 8 coupling blocks with fully-connected networks, denoted as \textit{subnetworks}, as scale and translation functions.
Each subnetwork upscales its input to 1024 dimensions with a fully-connected layer.
This is followed by a ReLU and a second fully-connected layer with dimension 48.
The condition encoding network $h_{\bm{\theta}}$ follows the same design with 256 and 56 as output dimensions of the fully-connected layers. 
We set the clamping parameter inside the coupling blocks to $\alpha = 2.0$.
For $\mathcal{L}_{\text{MMD}}$, we follow \cite{ArdizzoneINN} and employ a mixture of inverse multiquadratics kernels
\begin{equation}
\varphi^{im}_{\mathcal{S}}(\bm{v}, \hat{\bm{v}})=\sum_{b \in \mathcal{S}} \frac{b}{b+\left\|\bm{v}-\hat{\bm{v}}\right\|^2}
\end{equation}
with bandwidth parameters $\mathcal{S} = \{0.0025, 0.04, 0.81\}$.

\textbf{Training:}
The overall network is trained for 155 epochs using Adam \cite{Adam_KingmaB14} with an initial learning rate of $1 \cdot 10^{-4}$ and momentum values $\beta_1 = 0.5$ and $\beta_2 = 0.9$.
The learning rate is halved after 150 epochs, and a batch size of 64 is used.
To improve optimization stability, we clip the gradients in the range $[-15, 15]$.
During training, the covariance matrices are computed from $L = 200$ 3D pose hypotheses and the standard deviation threshold for the masking of $\mathcal{L}_{\text{HM}}$~(Eq.~\ref{eq:masking}) is set to $\sigma_t = 1.05 \cdot  \sigma_{\text{gt}} = 2.1$.
The weights of the different losses are set to $\lambda_{\text{MMD}} = 10$, $\lambda_{\text{det}} = \lambda_{\text{MB}} = 4$ and  $\lambda_{\text{HM}} = 750$, and the number of best hypotheses selected in $\mathcal{L}_{\text{MB}}$ to $k = 5$.
Since we estimate 3D poses in metric scale, there needs to be a conversion factor defined to relate between covariance matrices from pose hypotheses and from heatmaps.
We empirically found a good conversion factor to be $1$\,px $\mathrel{\widehat{=}} 10$\,mm.

\begin{table*}
	\begin{center}
		\resizebox{\linewidth}{!}{%
			\begin{tabular}{  c  c  c  c  c  c  c  c  c  c  c  c  c  c  c  c  c}
				\hline
				Protocol 1 (MPJPE) & Direct. & Disc. & Eat & Greet & Phone & Photo & Pose & Purch. & Sit & SitD & Smoke & Wait & WalkD & Walk & WalkT & Avg. \\ \hline 
				Martinez \etal~\cite{martinez_2017_3dbaseline} ($M=1$)& 51.8 & 56.2 & 58.1 & 59.0 & 69.5 & 78.4 & 55.2 & 58.1 & 74.0 & 94.6 & 62.3 & 59.1 & 65.1 & 49.5 & 52.4 & 62.9\\ \hdashline
				Li \etal~\cite{li2020weakly} ($M=10$) & {62.0} & {69.7} & {64.3} & {73.6} & {75.1} & {84.8} & {68.7} & {75.0} & {81.2} & {104.3} & {70.2} & {72.0} & {75.0} & {67.0} & {69.0} & {73.9}\\   
				Li \etal~\cite{Li_2019_CVPR} ($M=5$) & 43.8 & 48.6 & 49.1 & 49.8 & 57.6 & 61.5 & 45.9 & 48.3 & 62.0 & 73.4 & 54.8 & 50.6 & 56.0 & 43.4 & 45.5 & 52.7\\
				Oikarinen \etal~\cite{oikarinen2020graphmdn} ($M=200$) & 40.0 & \underline{43.2} & \underline{41.0} & \underline{43.4} & \underline{50.0} & \underline{53.6} & {40.1} & \underline{41.4} & \underline{52.6} & 67.3 & \underline{48.1} & $\bm{44.2}$ & $\bm{44.9}$ & 39.5 & \underline{40.2} & \underline{46.2}\\
				Sharma \etal~\cite{Sharma_2019_ICCV} ($M=200$) & $\bm{37.8}$ & \underline{43.2} & 43.0 & 44.3 & 51.1 & {57.0} & $\bm{39.7}$ & {43.0} & {56.3} & \underline{64.0} & \underline{48.1} & \underline{45.4} & {50.4} & \underline{37.9} & $\bm{39.9}$ & 46.8\\ \hdashline
				Ours ($\bm{z}_0$) ($M=1$) & {52.4} & {60.2} & {57.8} & {57.4} & {65.7} & {74.1} & {56.2} & {59.1} &69.3 & {78.0} & {61.2} & {63.7} & {67.0} & {50.0} & {54.9} & {61.8} \\ 
				Ours ($M=200$) & \underline{38.5} & $\bm{42.5}$ & $\bm{39.9}$ & $\bm{41.7}$ & $\bm{46.5}$ & $\bm{51.6}$ & \underline{39.9} & $\bm{40.8}$ & $\bm{49.5}$ & $\bm{56.8}$ & $\bm{45.3}$ & {46.4} & \underline{46.8} & $\bm{37.8}$ & {40.4} & $\bm{44.3}$\\ 
				\hline\hline
				Protocol 2 (PMPJPE) & Direct. & Disc. & Eat & Greet & Phone & Photo & Pose & Purch. & Sit & SitD & Smoke & Wait & WalkD & Walk & WalkT & Avg. \\ \hline 
				Martinez \etal~\cite{martinez_2017_3dbaseline} ($M=1$) & 39.5 & 43.2 & 46.4 & 47.0 & 51.0 & 56.0 & 41.4 & 40.6 & 56.5 & 69.4 & 49.2 & 45.0 & 49.5 & 38.0 & 43.1 & 47.7\\ \hdashline
				Li \etal~\cite{li2020weakly} ($M=10$) & {38.5} & {41.7} & {39.6} & {45.2} & {45.8} & {46.5} & {37.8} & {42.7} & {52.4} & {62.9} & {45.3} & {40.9} & {45.3} & {38.6} & {38.4} & {44.3}\\   
				Li \etal~\cite{Li_2019_CVPR} ($M=5$) & {35.5} & {39.8} & {41.3} & {42.3} & {46.0} & {48.9} & {36.9} & {37.3} & {51.0} & {60.6} & {44.9} & {40.2} & {44.1} & {33.1} & {36.9} & {42.6}\\    
				Oikarinen \etal~\cite{oikarinen2020graphmdn} ($M=200$) & 30.8 & 34.7 & \underline{33.6} & \underline{34.2} & \underline{39.6} & \underline{42.2} & \underline{31.0} & 31.9 & \underline{42.9} & 53.5 & \underline{38.1} & \underline{34.1} & \underline{38.0} & \underline{29.6} & \underline{31.1}& \underline{36.3}\\
				*Sharma \etal~\cite{Sharma_2019_ICCV} ($M=200$) & \underline{30.6} & \underline{34.6} & 35.7 & 36.4 & 41.2 & 43.6 & 31.8 & \underline{31.5} & 46.2 & \underline{49.7} & 39.7 & 35.8 & 39.6 & 29.7& 32.8 & 37.3\\ \hdashline
				Ours ($\bm{z}_0$) ($M=1$) & 37.8 & 41.7 & 42.1 & 41.8 & 46.5 & 50.2 & 38.0 & 39.2 & 51.7 & 61.8 & 45.4 & 42.6 & 45.7 & 33.7 & 38.5 & 43.8\\   
				Ours ($M=200$) & $\bm{27.9}$ & $\bm{31.4}$ & $\bm{29.7}$ & $\bm{30.2}$ & $\bm{34.9}$ & $\bm{37.1}$ & $\bm{27.3}$ & $\bm{28.2}$ & $\bm{39.0}$ & $\bm{46.1}$ & $\bm{34.2}$ & $\bm{32.3}$ & $\bm{33.6}$ & $\bm{26.1}$ & $\bm{27.5}$ & $\bm{32.4}$\\ \hline
			\end{tabular}%
		}
		\caption{Detailed results of MPJPE in millimetres on Human3.6M under Protocol~1 (no rigid alignment) and Protocol~2 (rigid alignment).
			Our model achieves state-of-the-art results, outperforming all other methods in nearly every activity.
			All scores are taken from the referenced papers, except the row marked with * which is computed using the publicly available official code and model from \cite{Sharma_2019_ICCV}. The number of samples estimated by the respective approaches is denoted as $M$.}
		\vspace{-2.0em}
		\label{table:protocolI}
	\end{center}
\end{table*}

\subsection{Evaluation on Human3.6M}
We follow previous works and report metrics for the best 3D pose hypothesis generated by our network. 
This is especially reasonable for ambiguous examples, where multiple diverse 3D poses form a correct solution for the 3D pose reconstruction.
Therefore, instead of validating whether predictions are equal to a specific solution, we evaluate if that specific solution is contained in the set of predictions.
Additionally, we show results for 3D poses generated with an all-zero latent vector $\bm{z}_0$.
Since we sample $\bm{z}$ from $\mathcal{N}(0, I)$ during training, such poses are approximately the highest likelihood solutions.
Following \cite{Sharma_2019_ICCV}, we produce $M = 200$ hypotheses for each 2D input. 
The results of our approach and other state-of-the-art methods are shown in Table~\ref{table:protocolI}. 
We outperform every competitor and achieve a clear improvement of $4.1\%$ and $10.7\%$ over the previous best scores under Protocol~1 and Protocol~2.
Note that Li \etal \cite{Li_2019_CVPR} only show detailed results for $M = 5$, but state that their model performance does not significantly improve when increasing $M$.
We generated the numbers for \cite{Sharma_2019_ICCV} under Protocol~2 (row marked with *) using their publicly available model, code and data, because they only report scores for the PMPJPE on subject 11.
Outperforming the single prediction baseline of \cite{martinez_2017_3dbaseline} with $\bm{z}_0$ generated poses (\ie $M = 1$) shows that our model is additionally able to give strong single predictions.

To evaluate the performance on highly ambiguous examples, we compute results for the challenging subset H36MA. 
We use the publicly available code from \cite{Sharma_2019_ICCV} and \cite{Li_2019_CVPR} to compare with their approaches.
As is shown in Table~\ref{table:hardsubset}, we outperform both competitors significantly and by a larger margin than on the whole test set.
This emphasizes the ability of our model to generate diverse hypotheses for highly ambiguous examples.
We argue that the CPS is especially meaningful in this setting, since high individual joint errors that often occur for challenging poses cannot be averaged out as in \eg MPJPE or PCK.

\begin{table}
	\begin{center}
		\resizebox{\linewidth}{!}{%
			\begin{tabular}{  c  c  c  c  c}
				\hline
				Method & MPJPE$\downarrow$ & PMPJPE$\downarrow$ & PCK$\uparrow$ & CPS$\uparrow$ \\ \hline 
				Li \etal~\cite{Li_2019_CVPR} & 81.1 & 66.0 & 85.7 & 119.9\\ 
				Sharma \etal~\cite{Sharma_2019_ICCV} & \underline{78.3} & \underline{61.1} & \underline{88.5} &  \underline{136.4} \\
				Ours  & $\bm{71.0}$ & $\bm{54.2}$ & $\bm{93.4}$ & $\bm{171.0}$\\ \hline
			\end{tabular}%
		}
		\caption{Evaluation results on the subset H36MA containing highly ambiguous examples.
			For each metric, the best hypothesis score is reported.}
		\vspace{-1.7em}
		\label{table:hardsubset}
	\end{center}
\end{table}

\subsection{Transfer to MPI-INF-3DHP}
To assess the generalization capability of our model, we evaluate on MPI-INF-3DHP.
Note that neither the 2D detector nor the normalizing flow is trained on this dataset. 
The results are shown in Table~\ref{table:3dhp}.
Even though \cite{Li_2019_CVPR} use the ground-truth 2D joints provided by the dataset, we clearly outperform them in all three settings.
We also achieve competitive performance compared to the weakly supervised approach from \cite{li2020weakly} that focuses on transfer learning.
Our strong results for outdoor scenes further emphasize the generalization capability to different settings.

\begin{table}
\begin{center}
\resizebox{\linewidth}{!}{%
    \begin{tabular}{c c  c  c  c}
    \hline
    Method & Studio GS & Studio no GS & Outdoor & All PCK \\ \hline 
    Li \etal~\cite{li2020weakly} & $\bm{86.9}$ & $\bm{86.6}$ & \underline{79.3} & $\bm{85.0}$\\
    Li \etal~\cite{Li_2019_CVPR} & 70.1 & 68.2 & 66.6 & 67.9\\
    Ours & \underline{86.6} & \underline{82.8} &  $\bm{82.5}$ & \underline{84.3}\\ \hline
    \end{tabular}%
    }
    \caption{Quantitative results on MPI-INF-3DHP. 
    We outperform the approach from \cite{Li_2019_CVPR} by a large margin which even uses ground truth 2D joint positions. Note that \cite{li2020weakly} is trained weakly supervised and therefore specifically built for transfer learning. 
    However, we still achieve on par results and even outperform them in the challenging outdoor sequences.}
    \vspace{-1.5em}
    \label{table:3dhp}
\end{center}
\end{table}

\subsection{Sample Diversity}

\begin{figure*}
\begin{center}
\includegraphics[width=0.95\linewidth]{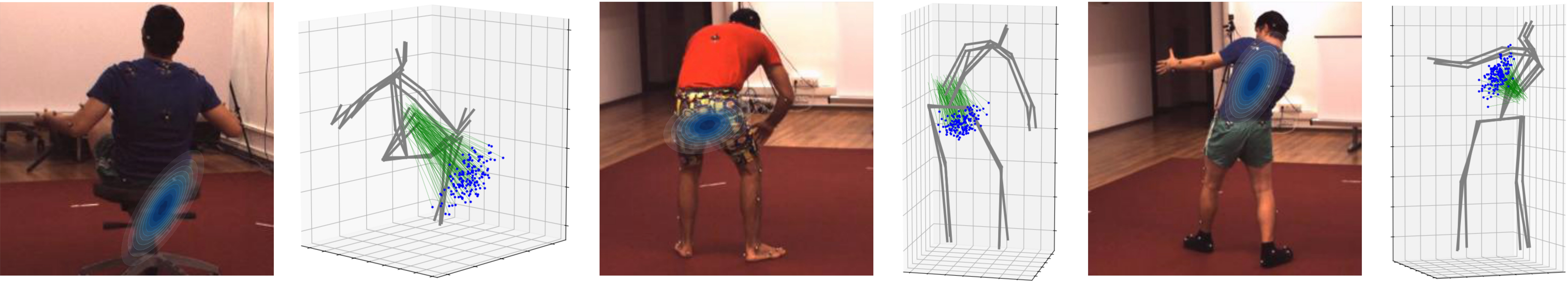}
\caption{The uncertainties of the 2D detector are successfully reflected in the 3D pose hypotheses. For visualization purposes, we only show the fitted Gaussian and a high number of hypotheses for the joint with highest uncertainty.}
\vspace{-1.9em}
\label{fig:hm_div}
\end{center}
\end{figure*}

\textbf{Heatmap Variance:} We visually inspect the distribution of generated joint locations and compare them with the corresponding fitted Gaussians in Fig.~\ref{fig:hm_div}. For visualization purposes, only three hypotheses are shown for all joints except the one with the highest uncertainty. As can be seen, the uncertainties of the 2D detector are reflected in the 3D hypotheses.

\begin{figure}
\begin{center}
\includegraphics[width=0.80\linewidth]{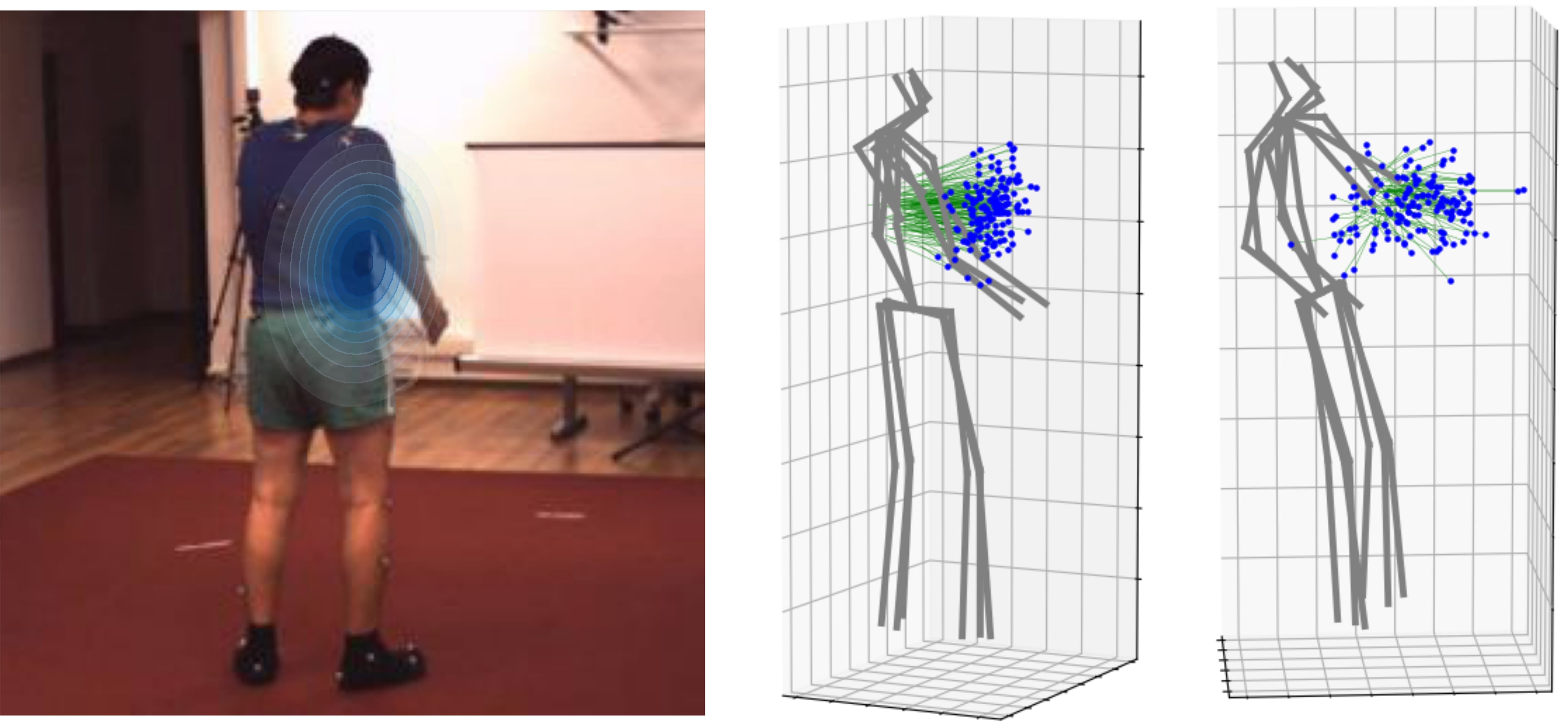}
\caption{Depth ambiguities can be modeled together with the uncertainties of the 2D detector.}
\vspace{-1.5em}
\label{fig:depth_var}
\end{center}
\end{figure}

\textbf{Depth Ambiguities:}
Even though variance in the depth direction is not explicitly optimized, our model learns to generate feasible hypotheses with varying depth.
In fact, the standard deviation of the hypotheses averaged over all joints in the test set of Human3.6M is highest in the depth direction with $42.4$\,mm, compared to $18.3$\,mm and $17.3$\,mm in the $x$- and $y$-directions.
Ankle, elbow and wrist joints account for the highest amount of variance.
A visual example of meaningful depth diversity is given in Fig.~\ref{fig:depth_var}.

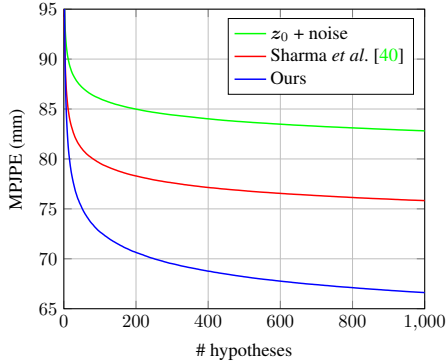
\begin{figure}
	\begin{center}
		\begin{tikzpicture}[scale=0.7]
		\begin{axis}[ 
		legend entries={$\bm{z}_0$ + noise,
		    Sharma \etal~\cite{Sharma_2019_ICCV},
		    Ours,},
		legend cell align={left},
		xmax=1000,
		xmin=0,
		ymax=95,
		ymin=65,
		legend pos=north east,
		samples=100,
		xtick={0,200,...,1000},
		ytick={65,70,75,...,95},
		grid=major,
		xlabel={\text{\# hypotheses}},
		y label style={at={(axis description cs:0.09,.5)},anchor=south},
		ylabel={MPJPE (mm)}
		] 
		\addplot[color=green, style=thick] table [x=n_hypos, y=z0_gauss, col sep=comma] {error_list_plot.csv};
		\addplot[color=red, style=thick] table [x=n_hypos, y=sharma, col sep=comma] {error_list_plot.csv};
	\addplot[color=blue, style=thick] table [x=n_hypos, y=ours, col sep=comma] {error_list_plot.csv};

	\end{axis}
		\end{tikzpicture}
\caption{
Evaluation results on the subset H36MA for an increasing number of hypotheses.
Our model further improves and enlarges the gap to \cite{Sharma_2019_ICCV} and to a noise baseline.
}
\vspace{-1.5em}
		\label{fig:hypoVsError}
	\end{center}
\end{figure}

\textbf{Sample Set Size and Noise Baseline:} In Fig.~\ref{fig:hypoVsError}, we plot the MPJPE on the subset H36MA with increasing number of samples.
The best hypothesis performance of our model continues to improve significantly, further enlarging the gap to \cite{Sharma_2019_ICCV}.
To validate that our approach is superior to directly sampling from the fitted Gaussians, we also plot results for a sampling baseline.
The baseline is constructed by adding noise sampled from the fitted Gaussians to each joint of the $\bm{z}_0$ predictions.
A constant Gaussian for the depth dimension is assumed.
Evidently, the performance of this baseline saturates earlier and at a higher error.

\subsection{Ablation Studies} \label{sec:ablations}
To quantify the influence of our proposed components and loss functions, we remove them individually and show the results in Table~\ref{table:ablation}.
As can be seen, the removal of each component leads to a degradation in performance.
When removing the heatmap condition, a large drop in the CPS can be observed.
This shows that some individual joints cannot be reconstructed without the uncertainty information of the 2D detector.
Providing the condition alone already leads to a significant improvement of the CPS, indicating that the network can automatically leverage the information to model ambiguities.
Adding $\mathcal{L}_{\text{HM}}$ further improves all metrics.
The importance of the discriminator becomes especially evident when considering the worst hypothesis error instead of the best.
For example, Protocol 2 computed for the worst hypothesis deteriorates from $86.8$\,mm to $284.1$\,mm without the discriminator.
Thus, the adversarial training procedure ensures the feasibility of the generated poses.
Table~\ref{table:ablation} also shows the influence of the number of best hypotheses $k$ selected for computing $\mathcal{L}_{\text{MB}}$. Note that $\mathcal{L}_{\text{MB}}$ with $k=1$ is equivalent to the typical \textit{best-of-M} loss.

\begin{table}
\begin{center}
\resizebox{0.9\linewidth}{!}{
    \begin{tabular}{  c  c  c  c  c}
    \hline
    Method & MPJPE$\downarrow$ & PMPJPE$\downarrow$ & PCK$\uparrow$ & CPS$\uparrow$ \\ \hline 
    w/o condition & {71.7} & 57.2 & 91.2 & 137.6 \\
    w/o $\mathcal{L}_{\text{HM}}$ & 72.4 & {56.2} & 92.2 & 157.3\\ 
    w/o $\mathcal{L}_{\text{gen}}$ & 73.6 & 58.2 & {92.5} &  {165.5}\\ \hdashline
     w/o $\mathcal{L}_{\text{MB}}$ & 76.0 & 58.5 & 91.6 & 161.4  \\ 
     $\mathcal{L}_{\text{MB}} (k=1)$ & 71.8 & \underline{54.9} & 92.6 &  167.4  \\
    $\mathcal{L}_{\text{MB}} (k=10)$ & $\bm{70.7}$ & \underline{54.9} & \underline{93.3} &  \underline{168.3}\\
    $\mathcal{L}_{\text{MB}} (k=50)$ & \underline{71.0} & 55.2 & 93.1 &  168.0\\ \hdashline
    Ours (Full) & \underline{71.0} & $\bm{54.2}$ & $\bm{93.4}$ & $\bm{171.0}$\\ \hline
    \end{tabular}}
    \caption{Ablation studies on the subset H36MA.}
    \vspace{-1.5em}
    \label{table:ablation}
\end{center}
\end{table}

\section{Conclusion}
This paper presents a normalizing flow based method for the ambiguous inverse problem of 3D human pose estimation from 2D inputs.
We exploit the bijectivity of the normalizing flow by utilizing the known 3D to 2D projection during training.
By incorporating uncertainty information from the heatmaps of a 2D pose detector, valuable information is maintained which is discarded by previous approaches.
As demonstrated, the generated hypotheses reflect these uncertainties and additionally show meaningful diversity along the ambiguous depth of the joints.
Furthermore, the introduction of a 3D pose discriminator ensures the geometrical feasibility of the poses and a proposed generalization of the \textit{best-of-M} loss improves the performance. 
Experimental results show that our method outperforms all previous multi-hypotheses approaches in most metrics, especially on a challenging subset of Human3.6M containing highly ambiguous examples.
\vspace{-1.0em}
\small{\paragraph{Acknowledgements.}
This work has been supported by the Federal Ministry of Education and Research (BMBF), Germany, under the project LeibnizKILabor (grant no.\ 01DD20003), the Center for Digital Innovations (ZDIN) and the Deutsche Forschungsgemeinschaft (DFG) under Germany’s Excellence Strategy within the Cluster of Excellence PhoenixD (EXC 2122).}


\newpage
\clearpage
{\small
\bibliographystyle{ieee_fullname}
\bibliography{egbib}

\begin{thebibliography}{10}\itemsep=-1pt

\bibitem{Akhter_2015_CVPR}
Ijaz Akhter and Michael~J. Black.
\newblock Pose-conditioned joint angle limits for 3d human pose reconstruction.
\newblock In {\em The IEEE Conference on Computer Vision and Pattern
  Recognition (CVPR)}, 2015.

\bibitem{andriluka14cvpr}
Mykhaylo Andriluka, Leonid Pishchulin, Peter Gehler, and Bernt Schiele.
\newblock 2d human pose estimation: New benchmark and state of the art
  analysis.
\newblock In {\em The IEEE Conference on Computer Vision and Pattern
  Recognition (CVPR)}, 2014.

\bibitem{ArdizzoneINN}
Lynton Ardizzone, Jakob Kruse, Sebastian~J. Wirkert, Daniel Rahner, Eric~W.
  Pellegrini, Ralf~S. Klessen, Lena Maier{-}Hein, Carsten Rother, and Ullrich
  K{\"{o}}the.
\newblock Analyzing inverse problems with invertible neural networks.
\newblock {\em International Conference on Learning Representations (ICLR)},
  2019.

\bibitem{Ardizzone2019a}
Lynton Ardizzone, Carsten Lüth, Jakob Kruse, Carsten Rother, and Ullrich
  Köthe.
\newblock Guided image generation with conditional invertible neural networks.
\newblock {\em arXiv preprint arXiv:1907.02392}, 2019.

\bibitem{multi_bodies_biggs2020}
Benjamin Biggs, Sébastien Ehrhadt, Hanbyul Joo, Benjamin Graham, Andrea
  Vedaldi, and David Novotny.
\newblock 3d multi-bodies: Fitting sets of plausible 3d human models to
  ambiguous image data.
\newblock In {\em Advances in Neural Information Processing Systems (NeurIPS)},
  2020.

\bibitem{370fbeadb5584ba9ab2938431fc4f140}
{Christopher M.} Bishop.
\newblock Mixture density networks.
\newblock Technical report, Aston University, 1994.

\bibitem{BogoECCV2016}
Federica Bogo, Angjoo Kanazawa, Christoph Lassner, Peter Gehler, Javier Romero,
  and Michael~J. Black.
\newblock Keep it {SMPL}: Automatic estimation of {3D} human pose and shape
  from a single image.
\newblock In {\em European Conference on Computer Vision (ECCV)}, 2016.

\bibitem{Chen_2017_CVPR}
Ching-Hang Chen and Deva Ramanan.
\newblock 3d human pose estimation = 2d pose estimation + matching.
\newblock In {\em The IEEE Conference on Computer Vision and Pattern
  Recognition (CVPR)}, 2017.

\bibitem{Ci_2019_ICCV}
Hai Ci, Chunyu Wang, Xiaoxuan Ma, and Yizhou Wang.
\newblock Optimizing network structure for 3d human pose estimation.
\newblock In {\em Proceedings of the IEEE International Conference on Computer
  Vision (ICCV)}, 2019.

\bibitem{DBLP:journals/corr/DinhKB14}
Laurent Dinh, David Krueger, and Yoshua Bengio.
\newblock {NICE:} non-linear independent components estimation.
\newblock In {\em International Conference on Learning Representations (ICLR)},
  2015.

\bibitem{DBLP:conf/iclr/DinhSB17}
Laurent Dinh, Jascha Sohl{-}Dickstein, and Samy Bengio.
\newblock Density estimation using real {NVP}.
\newblock In {\em International Conference on Learning Representations (ICLR)},
  2017.

\bibitem{fang2018learning}
Haoshu Fang, Yuanlu Xu, Wenguan Wang, Xiaobai Liu, and Song-Chun Zhu.
\newblock Learning pose grammar to encode human body configuration for 3d pose
  estimation.
\newblock In {\em Proceedings of the AAAI Conference on Artificial
  Intelligence}, 2018.

\bibitem{pmlr-v37-germain15}
Mathieu Germain, Karol Gregor, Iain Murray, and Hugo Larochelle.
\newblock Made: Masked autoencoder for distribution estimation.
\newblock In {\em Proceedings of the International Conference on Machine
  Learning (ICML)}, 2015.

\bibitem{JMLR:v13:gretton12a}
Arthur Gretton, Karsten~M. Borgwardt, Malte~J. Rasch, Bernhard Sch{{\"o}}lkopf,
  and Alexander Smola.
\newblock A kernel two-sample test.
\newblock {\em Journal of Machine Learning Research (JMLR)}, 13(25), 2012.

\bibitem{NIPS2017_wgangp}
Ishaan Gulrajani, Faruk Ahmed, Martin Arjovsky, Vincent Dumoulin, and Aaron~C
  Courville.
\newblock Improved training of wasserstein gans.
\newblock In {\em Advances in Neural Information Processing Systems (NeurIPS)},
  2017.

\bibitem{NIPS2012_cfbce4c1}
Abner Guzm\'{a}n-rivera, Dhruv Batra, and Pushmeet Kohli.
\newblock Multiple choice learning: Learning to produce multiple structured
  outputs.
\newblock In {\em Advances in Neural Information Processing Systems (NeurIPS)},
  2012.

\bibitem{inthewild3d_2019}
Ikhsanul Habibie, Weipeng Xu, Dushyant Mehta, Gerard Pons-Moll, and Christian
  Theobalt.
\newblock In the wild human pose estimation using explicit 2d features and
  intermediate 3d representations.
\newblock In {\em The IEEE Conference on Computer Vision and Pattern
  Recognition (CVPR)}, 2019.

\bibitem{Hossain2018ECCV}
Mir Rayat~Imtiaz Hossain and James~J. Little.
\newblock Exploiting temporal information for 3d pose estimation.
\newblock {\em European Conference on Computer Vision (ECCV)}, 2018.

\bibitem{h36m_pami}
Catalin Ionescu, Dragos Papava, Vlad Olaru, and Cristian Sminchisescu.
\newblock Human3.6m: Large scale datasets and predictive methods for 3d human
  sensing in natural environments.
\newblock {\em IEEE Transactions on Pattern Analysis and Machine Intelligence
  (PAMI)}, 36(7), 2014.

\bibitem{Jahangiri2017GeneratingMD}
Ehsan Jahangiri and Alan~L. Yuille.
\newblock Generating multiple diverse hypotheses for human 3d pose consistent
  with 2d joint detections.
\newblock {\em International Conference on Computer Vision Workshops (ICCVW)},
  2017.

\bibitem{KanazawaCVPR18}
Angjoo Kanazawa, Michael~J. Black, David~W. Jacobs, and Jitendra Malik.
\newblock End-to-end recovery of human shape and pose.
\newblock In {\em The IEEE Conference on Computer Vision and Pattern
  Recognition (CVPR)}, 2018.

\bibitem{Adam_KingmaB14}
Diederik~P. Kingma and Jimmy Ba.
\newblock Adam: {A} method for stochastic optimization.
\newblock In {\em International Conference on Learning Representations (ICLR)},
  2015.

\bibitem{NIPS2016_6581}
Durk~P Kingma, Tim Salimans, Rafal Jozefowicz, Xi Chen, Ilya Sutskever, and Max
  Welling.
\newblock Improved variational inference with inverse autoregressive flow.
\newblock In {\em Advances in Neural Information Processing Systems (NeurIPS)},
  2016.

\bibitem{kob2020NF}
Ivan Kobyzev, Simon Prince, and Marcus Brubaker.
\newblock Normalizing flows: An introduction and review of current methods.
\newblock {\em TPAMI}, 2020.

\bibitem{kocabas2019vibe}
Muhammed Kocabas, Nikos Athanasiou, and Michael~J. Black.
\newblock Vibe: Video inference for human body pose and shape estimation.
\newblock In {\em The IEEE Conference on Computer Vision and Pattern
  Recognition (CVPR)}, June 2020.

\bibitem{kolotouros2019spin}
Nikos Kolotouros, Georgios Pavlakos, Michael~J. Black, and Kostas Daniilidis.
\newblock Learning to reconstruct 3d human pose and shape via model-fitting in
  the loop.
\newblock In {\em Proceedings of the IEEE International Conference on Computer
  Vision (ICCV)}, 2019.

\bibitem{LeeCo04}
Mun~Wai Lee and Isaac Cohen.
\newblock Proposal maps driven mcmc for estimating human body pose in static
  images.
\newblock In {\em The IEEE Conference on Computer Vision and Pattern
  Recognition (CVPR)}, 2004.

\bibitem{Li_2019_CVPR}
Chen Li and Gim~Hee Lee.
\newblock Generating multiple hypotheses for 3d human pose estimation with
  mixture density network.
\newblock In {\em The IEEE Conference on Computer Vision and Pattern
  Recognition (CVPR)}, 2019.

\bibitem{li2020weakly}
Chen Li and Gim~Hee Lee.
\newblock Weakly supervised generative network for multiple 3d human pose
  hypotheses.
\newblock {\em British Machine Vision Conference (BMVC)}, 2020.

\bibitem{li21hybrik}
Jiefeng Li, Chao Xu, Zhicun Chen, Siyuan Bian, Lixin Yang, and Cewu Lu.
\newblock Hybrik: A hybrid analytical-neural inverse kinematics solution for 3d
  human pose and shape estimation.
\newblock In {\em CVPR}, 2021.

\bibitem{Li_2020_CVPR}
Shichao Li, Lei Ke, Kevin Pratama, Yu-Wing Tai, Chi-Keung Tang, and Kwang-Ting
  Cheng.
\newblock Cascaded deep monocular 3d human pose estimation with evolutionary
  training data.
\newblock In {\em The IEEE Conference on Computer Vision and Pattern
  Recognition (CVPR)}, 2020.

\bibitem{martinez_2017_3dbaseline}
Julieta Martinez, Rayat Hossain, Javier Romero, and James~J. Little.
\newblock A simple yet effective baseline for 3d human pose estimation.
\newblock In {\em Proceedings of the IEEE International Conference on Computer
  Vision (ICCV)}, 2017.

\bibitem{mono-3dhp2017}
Dushyant Mehta, Helge Rhodin, Dan Casas, Pascal Fua, Oleksandr Sotnychenko,
  Weipeng Xu, and Christian Theobalt.
\newblock Monocular 3d human pose estimation in the wild using improved cnn
  supervision.
\newblock In {\em International Conference on 3D Vision (3DV)}, 2017.

\bibitem{Moreno-Noguer_2017_CVPR}
Francesc Moreno-Noguer.
\newblock 3d human pose estimation from a single image via distance matrix
  regression.
\newblock In {\em The IEEE Conference on Computer Vision and Pattern
  Recognition (CVPR)}, 2017.

\bibitem{oikarinen2020graphmdn}
Tuomas~P. Oikarinen, Daniel~C. Hannah, and Sohrob Kazerounian.
\newblock Graphmdn: Leveraging graph structure and deep learning to solve
  inverse problems.
\newblock {\em arXiv preprint arXiv:2010.13668}, 2020.

\bibitem{NIPS2017_6828}
George Papamakarios, Theo Pavlakou, and Iain Murray.
\newblock Masked autoregressive flow for density estimation.
\newblock In {\em Advances in Neural Information Processing Systems (NeurIPS)},
  2017.

\bibitem{pavlaCVPR18}
Georgios Pavlakos, Luyang Zhu, Xiaowei Zhou, and Kostas Daniilidis.
\newblock Vibe: Video inference for human body pose and shape estimation.
\newblock In {\em The IEEE Conference on Computer Vision and Pattern
  Recognition (CVPR)}, June 2018.

\bibitem{pmlr-v37-rezende15}
Danilo Rezende and Shakir Mohamed.
\newblock Variational inference with normalizing flows.
\newblock Proceedings of Machine Learning Research (PMLR), 2015.

\bibitem{RudWan2021}
Marco Rudolph, Bastian Wandt, and Bodo Rosenhahn.
\newblock Same same but differnet: Semi-supervised defect detection with
  normalizing flows.
\newblock In {\em Winter Conference on Applications of Computer Vision (WACV)},
  2021.

\bibitem{Sharma_2019_ICCV}
Saurabh Sharma, Pavan~Teja Varigonda, Prashast Bindal, Abhishek Sharma, and
  Arjun Jain.
\newblock Monocular 3d human pose estimation by generation and ordinal ranking.
\newblock In {\em Proceedings of the IEEE International Conference on Computer
  Vision (ICCV)}, 2019.

\bibitem{PhysCapTOG2020}
Soshi Shimada, Vladislav Golyanik, Weipeng Xu, and Christian Theobalt.
\newblock Physcap: Physically plausible monocular 3d motion capture in real
  time.
\newblock {\em ACM Transactions on Graphics}, 39(6), 2020.

\bibitem{Simo12}
E. Simo-Serra, A. Ramisa, G. Alenyà, C. Torras, and F. Moreno-Noguer.
\newblock Single image 3d human pose estimation from noisy observations.
\newblock In {\em The IEEE Conference on Computer Vision and Pattern
  Recognition (CVPR)}, 2012.

\bibitem{SmiTri01}
Cristian Sminchisescu and Bill Triggs.
\newblock Covariance scaled sampling for monocular 3d body tracking.
\newblock In {\em The IEEE Conference on Computer Vision and Pattern
  Recognition (CVPR)}, 2001.

\bibitem{SmiTri03}
Cristian Sminchisescu and Bill Triggs.
\newblock Kinematic jump processes for monocular 3d human tracking.
\newblock In {\em The IEEE Conference on Computer Vision and Pattern
  Recognition (CVPR)}, 2003.

\bibitem{Sun_2019_CVPR}
Ke Sun, Bin Xiao, Dong Liu, and Jingdong Wang.
\newblock Deep high-resolution representation learning for human pose
  estimation.
\newblock In {\em The IEEE Conference on Computer Vision and Pattern
  Recognition (CVPR)}, 2019.

\bibitem{doi:10.1002/cpa.21423}
{Esteban G.} Tabak and Cristina~V. Turner.
\newblock A family of nonparametric density estimation algorithms.
\newblock {\em Communications on Pure and Applied Mathematics}, 66(2), 2013.

\bibitem{tabak2010}
{Esteban G.} Tabak and Eric Vanden-Eijnden.
\newblock Density estimation by dual ascent of the log-likelihood.
\newblock {\em Communications in Mathematical Sciences}, 8(1), 2010.

\bibitem{WanAck2018a}
Bastian Wandt, Hanno Ackermann, and Bodo Rosenhahn.
\newblock A kinematic chain space for monocular motion capture.
\newblock In {\em European Conference on Computer Vision Workshops (ECCVW)},
  2018.

\bibitem{Wandt2019RepNet}
Bastian Wandt and Bodo Rosenhahn.
\newblock Repnet: Weakly supervised training of an adversarial reprojection
  network for 3d human pose estimation.
\newblock In {\em The IEEE Conference on Computer Vision and Pattern
  Recognition (CVPR)}, 2019.

\bibitem{WanRud2021a}
Bastian Wandt, Marco Rudolph, Petrissa Zell, Helge Rhodin, and Bodo Rosenhahn.
\newblock Canonpose: Self-supervised monocular 3d human pose estimation in the
  wild.
\newblock In {\em The IEEE Conference on Computer Vision and Pattern
  Recognition (CVPR)}, 2021.

\bibitem{DBLP:journals/corr/abs-1905-07862}
Jue Wang, Shaoli Huang, Xinchao Wang, and Dacheng Tao.
\newblock Not all parts are created equal: 3d pose estimation by modelling
  bi-directional dependencies of body parts.
\newblock {\em Proceedings of the IEEE International Conference on Computer
  Vision (ICCV)}, 2019.

\bibitem{winkler2019learning}
Christina Winkler, Daniel Worrall, Emiel Hoogeboom, and Max Welling.
\newblock Learning likelihoods with conditional normalizing flows.
\newblock {\em arXiv preprint arXiv:1912.00042}, 2019.

\bibitem{xu2020ghum}
Hongyi Xu, Eduard~Gabriel Bazavan, Andrei Zanfir, William~T Freeman, Rahul
  Sukthankar, and Cristian Sminchisescu.
\newblock Ghum \& {G}huml: Generative 3d human shape and articulated pose
  models.
\newblock In {\em The IEEE Conference on Computer Vision and Pattern
  Recognition (CVPR)}, 2020.

\bibitem{Xu_2020_CVPR}
Jingwei Xu, Zhenbo Yu, Bingbing Ni, Jiancheng Yang, Xiaokang Yang, and Wenjun
  Zhang.
\newblock Deep kinematics analysis for monocular 3d human pose estimation.
\newblock In {\em The IEEE Conference on Computer Vision and Pattern
  Recognition (CVPR)}, 2020.

\bibitem{Zanfir2020WeaklyS3}
Andrei Zanfir, Eduard~Gabriel Bazavan, Hongyi Xu, Bill Freeman, Rahul
  Sukthankar, and Cristian Sminchisescu.
\newblock Weakly supervised 3d human pose and shape reconstruction with
  normalizing flows.
\newblock {\em European Conference on Computer Vision (ECCV)}, 2020.

\bibitem{zhaoCVPR19semantic}
Long Zhao, Xi Peng, Yu Tian, Mubbasir Kapadia, and Dimitris~N. Metaxas.
\newblock Semantic graph convolutional networks for 3d human pose regression.
\newblock In {\em The IEEE Conference on Computer Vision and Pattern
  Recognition (CVPR)}, 2019.

\end{thebibliography}
}


\renewcommand\thesection{\Alph{section}}
\renewcommand\thesubsection{\thesection.\arabic{subsection}}
\appendix

\section*{Appendix}

\section{Qualitative Evaluation}

\subsection{Condition Influence}
\begin{figure*}
\centering
  \includegraphics[width=1.0\linewidth]{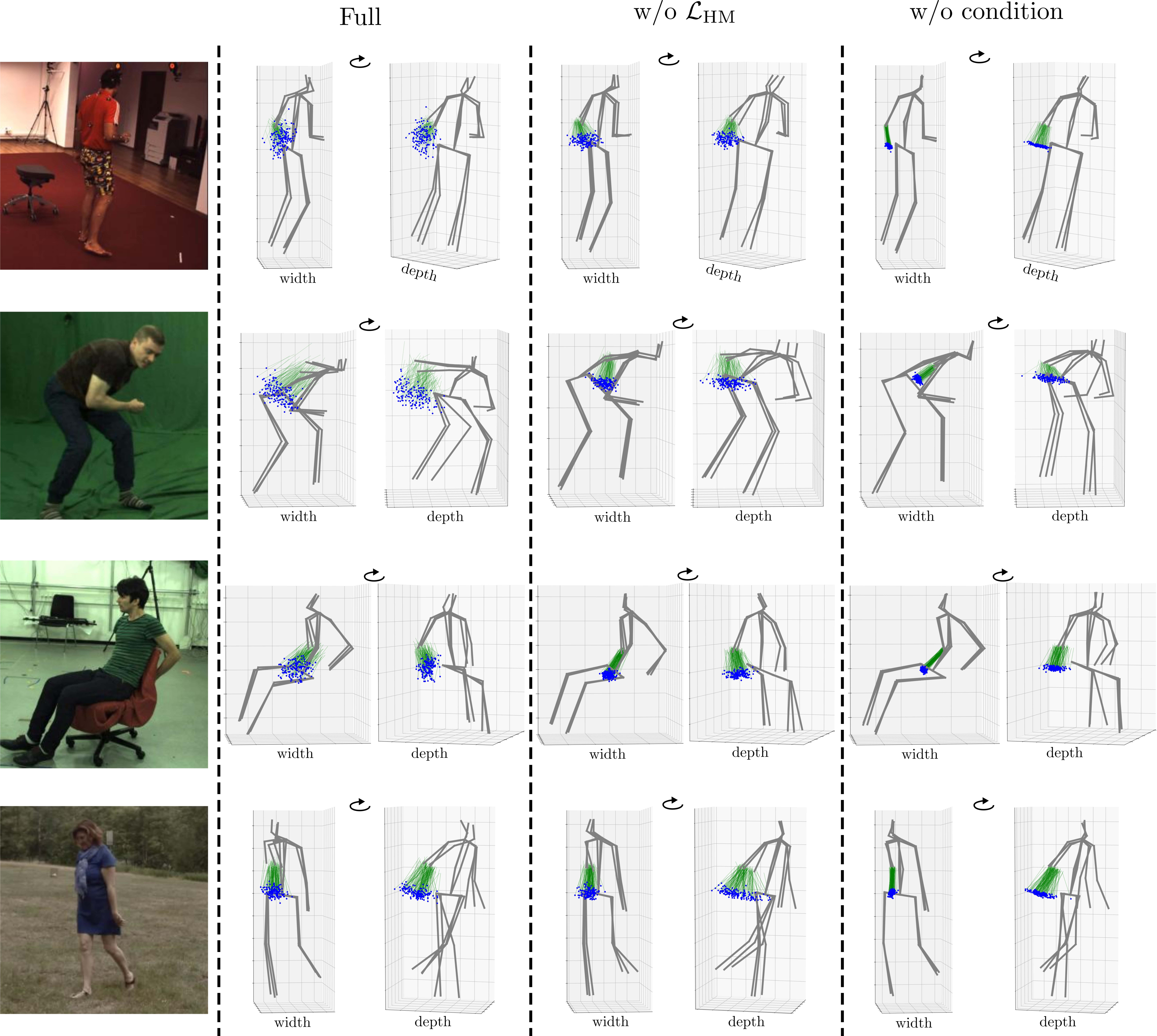} 
 \caption{Qualitative results of our full model, model without $\mathcal{L}_{\text{HM}}$, and the model without condition.
 For visualization purposes, more than three hypotheses are shown only for the most ambiguous joint.}
\label{fig:cond}
\end{figure*}

To further show the influence of the heatmap condition and of the loss $\mathcal{L}_{\text{HM}}$ that forces the network to reflect the 2D detector uncertainty in the 3D hypotheses, we present several qualitative results in Fig.~\ref{fig:cond}.
Evidently, incorporating the heatmap condition alone already leads to meaningful diversity along the $x$- and $y$- directions.
Additionally optimizing $\mathcal{L}_{\text{HM}}$ further increases the meaningful diversity of the pose hypotheses such that the uncertainties of the 2D detector as well as the depth ambiguities are modeled best.

\subsection{Competitor Comparison}
\begin{figure*}
\centering
  \includegraphics[width=1.0\linewidth]{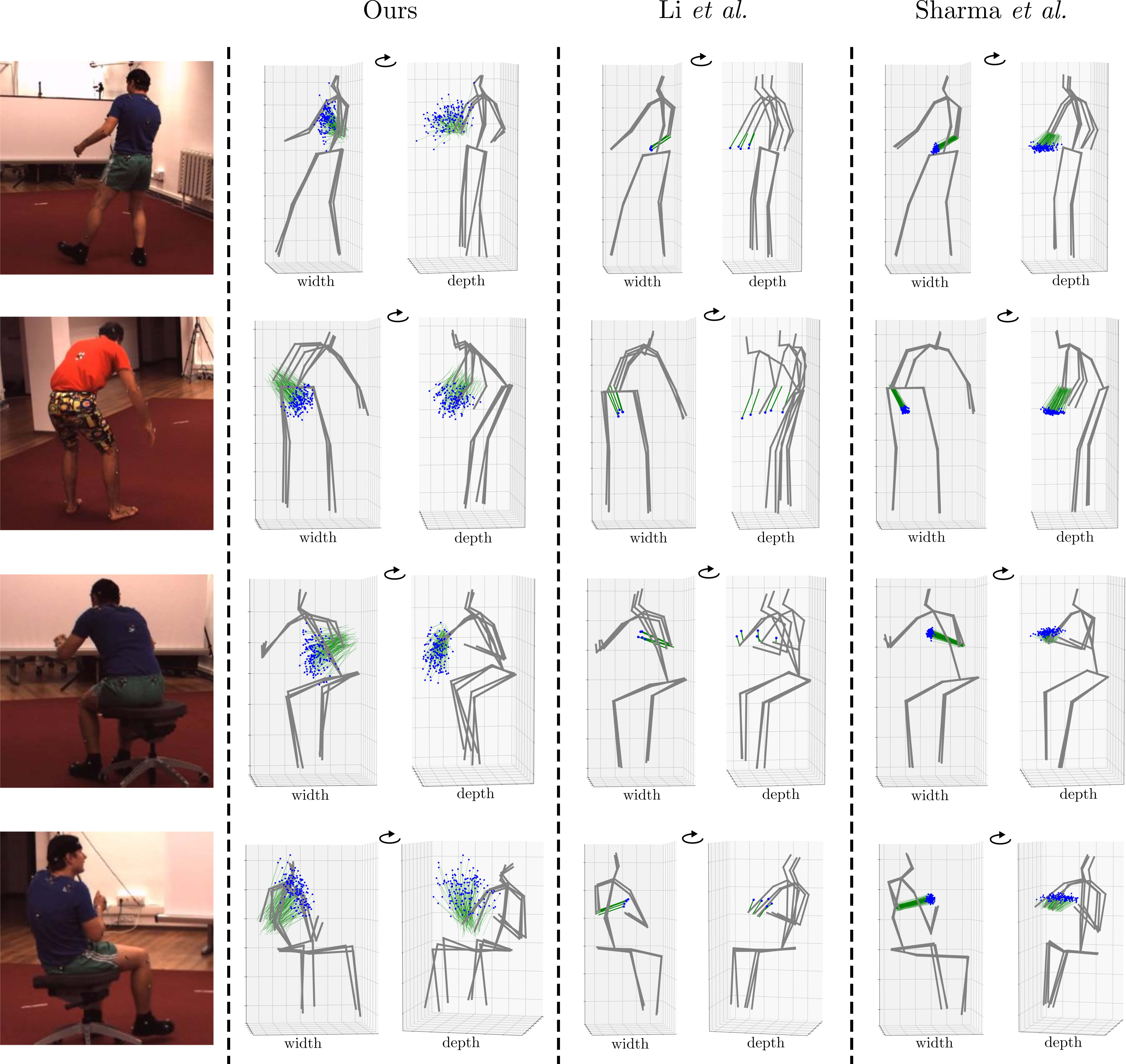} 
 \caption{Comparison with competing methods \cite{Li_2019_CVPR,Sharma_2019_ICCV}.
 For visualization purposes, more than three hypotheses are shown only for the most ambiguous joint.
 The model from Li \etal~\cite{Li_2019_CVPR} can only generate five pose hypotheses.
}
\label{fig:competitors}
\end{figure*}

In Fig.~\ref{fig:competitors}, we show additional qualitative results comparing our method with the competing methods \cite{Li_2019_CVPR,Sharma_2019_ICCV}.
As can be seen, our method achieves significantly higher diversity mainly for occluded joints.
The competing methods are unable to effectively model occlusions and uncertain detections.
They only achieve significant diversity along the ambiguous depth of the joints.

\subsection{High Confidence Detections}
\begin{figure*}
\centering
  \includegraphics[width=1.0\linewidth]{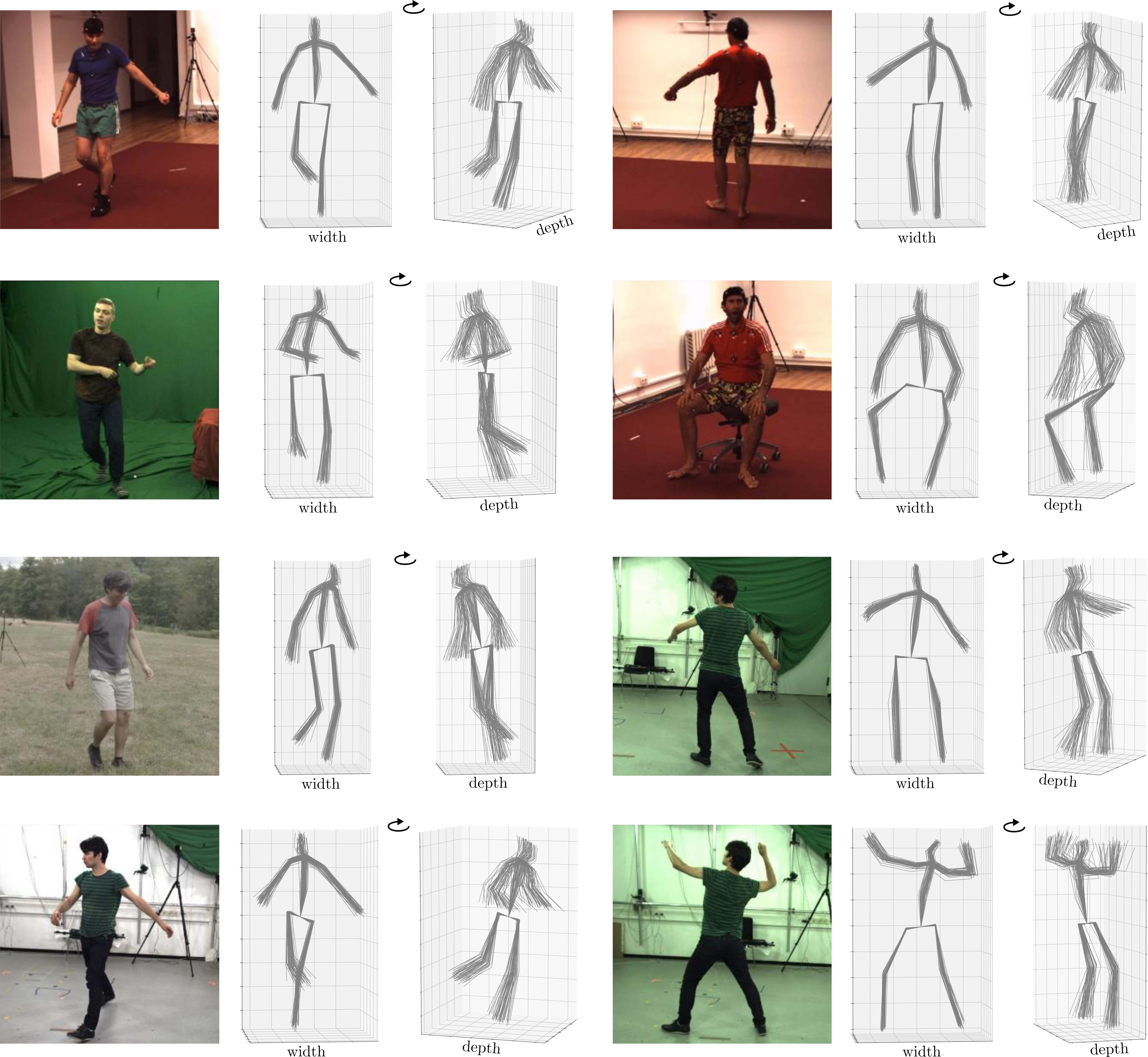} 
 \caption{Qualitative results for images from Human3.6M and MPI-INF-3DHP with low 2D detector uncertainty.
 For each image, 50 pose hypotheses are generated and shown from two perspectives.}
\label{fig:less_div}
\end{figure*}

If the 2D detector has a high degree of confidence for the 2D pose detection in a given image, then low variance in the generated 3D hypotheses along the $x$- and $y$- directions is expected.
To validate this, we show qualitative results for images from Human3.6M and MPI-INF-3DHP with low 2D detector uncertainty in Fig.~\ref{fig:less_div}. 
The generated hypotheses are shown from two perspectives such that diversity along the image and depth directions can be seen.
Evidently, the hypotheses vary only slightly along the image directions and thus are all consistent with the input image.
They show meaningful diversity along the ambiguous depth of the joints.

\section{Captured 2D Detector Uncertainty}
\begin{figure}
\begin{center}
\includegraphics[width=0.90\linewidth]{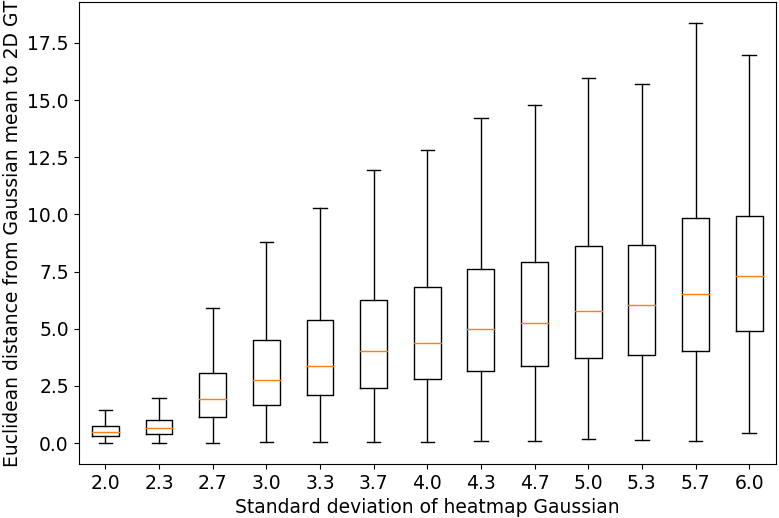}
\caption{Computed for all joints in the test split of Human3.6M.}
\label{fig:std_vs_error}
\end{center}
\end{figure}

In the following, we want to further verify that
fitting a Gaussian to the heatmap can capture the uncertainty of the 2D detector well.
Therefore, for each joint in the test split of Human3.6M, we show the mean of the standard deviations of the fitted Gaussian together with the 2D error in Fig.~\ref{fig:std_vs_error}.
As can be seen, the variances of the Gaussians correlate with the 2D error and thus are a good surrogate for the uncertainty of the 2D detector. 

\section{Performance Lower Number of Samples}
To assess the influence of the number of generated hypotheses and make our approach better comparable to Li~\etal~\cite{Li_2019_CVPR}, we evaluate on Human3.6M under Protocol~1 (MPJPE) and Protocol~2 (PMPJPE) for lower number of hypotheses in two different settings.
However, we want to emphasize that our main goal is to model the full posterior distribution, which requires a larger number of samples.
Instead of sampling from their model, Li \etal~\cite{Li_2019_CVPR} take the means of the Gaussian kernels as pose predictions.
Thus, for better comparison, we emulate this by running K-Means on our $M=200$ generated hypotheses.
Additionally, we compare the performance when \emph{sampling} from \cite{Li_2019_CVPR} in Table~\ref{table:h36m_testset} (rows marked with *).
We outperform them in almost every setting and metric.

\begin{table}
	\begin{center}
	\setlength{\tabcolsep}{2pt}
    \resizebox{\linewidth}{!}{
			\renewcommand{\arraystretch}{1.0}
			\begin{tabular}{  c  c  c  c | c  c  c  c}
				\hline
				Method & Hypo. & MPJPE$\downarrow$ & PMPJPE$\downarrow$ & Method & Hypo. & MPJPE$\downarrow$ & PMPJPE$\downarrow$\\ \hline 
				Li \cite{Li_2019_CVPR} & 5 & 52.7 & 42.6 & Ours (K-Means)  & 5 & ${53.2}$ & ${38.4}$\\ \hdashline
				*Li \cite{Li_2019_CVPR} & 5 & 74.9 & 63.3 & *Ours  & 5 & ${59.2}$ & ${42.3}$\\
				*Li \cite{Li_2019_CVPR} & 10 & 70.3 & 59.7 & *Ours  & 10 & ${55.0}$ & ${39.6}$\\
				*Li \cite{Li_2019_CVPR} & 200 & 59.6 & 50.2 & *Ours  & 200 & ${44.3}$ & ${32.4}$ \\ \hline
			\end{tabular}%
		}
		\caption{Results on Human3.6M under Protocol~1 (MPJPE) and Protocol~2 (PMPJPE). The scores for the rows marked with * are computed by sampling from the models. }
		\label{table:h36m_testset}
	\end{center}
\end{table}

\section{Inference Time}
For inference time measurements, we run the code with PyTorch 1.7.1 on a NVIDIA GeForce RTX 3090 (CUDA 11.4).
The majority of the inference time comes from the 2D detector ($32$\,ms) and the Gaussian fitting process (70\,ms).
Due to batch processing, generating multiple hypotheses brings nearly no overhead, with an inference time of $4.6$\,ms for a single and $5.1$\,ms for 1000 samples.

\end{document}